\documentstyle[12pt,mkstyle]{article}

\date{}
\newcommand{\Bem}[1]{}

\renewcommand{\baselinestretch}{0.95}

\date{}
\newcommand{\Sektion}[1]{\quad

\begin{center}
#1
\end{center}

}
\begin{document}


\begin{center}
REASONING FROM DATA IN THE MATHEMATICAL THEORY OF EVIDENCE\\
\quad \\
\quad \\
Mieczys{\l}aw A. K{\l}opotek\\
Institute of Computer Science, Polish Academy of Sciences\\
01-237 Warsaw, ul. Ordona 21, POLAND, e-mail: klopotek{@}plearn.bitnet
\end{center}

{
 \renewcommand{\baselinestretch}{1.2}
 {
\begin{abstract}

Mathematical Theory of Evidence (MTE) is known as a foundation
for reasoning when knowledge is expressed at various levels of
detail. Though much research effort has been committed to this
theory since its foundation, many questions remain open. One of
the most important open questions seems to be the relationship
between frequencies and the Mathematical Theory of Evidence.
The theory is blamed to leave frequencies outside (or aside of)
its framework. The seriousness of this accusation is obvious:
no experiment may be run to compare the performance of
MTE-based models of real world processes against real world
data.

In this paper we develop a frequentist model of the MTE
bringing to fall the above argument against MTE. We describe,
how to interpret data in terms of MTE belief functions, how to
reason from data about conditional belief functions, how to
generate a random sample out of a MTE model, how   to derive
MTE model from data and how to compare results of reasoning in
MTE model and reasoning from data.  

It is claimed in this paper that MTE is suitable to model some 
types of destructive processes
\end{abstract}

\Bem{   * Evolutionary Computation
   * Intelligent Information Systems
   * Knowledge Representation and Integration
   * ........ and Adaptive Systems
   * Logic for Artificial Intelligence
   * Methodologies (........, design, validation, performance evaluation).
}
 }
}


%
%
\Sektion{1. INTRODUCTION}
The Dempster-Shafer Theory  or the Mathematical Theory of Evidence (MTE) 
\cite{Shafer:76,Dempster:67} 
 shows one of possible ways of application of mathematical probability for
subjective evaluation and is intended to be a generalization of bayesian
theory of subjective probability \cite{Shafer:90ijar}. 

This theory offers a number of methodological advantages like: capability of
representing ignorance in a simple and direct way, compatibility with the
classical probability theory, compatibility of boolean logic and feasible
computational complexity  \cite{Ruspini:92ijar}. 

MTE may be applied for (1) representation of incomplete knowledge, (2) belief
updating, (3) and for combination of evidence  \cite{Provan:92}.
MTE covers the statistics of random sets and may be applied for representation
of incomplete statistical knowledge. Random set statistics is quite popular in
analysis of opinion polls whenever partial indecisiveness of respondents is
allowed \cite{Dubois:92}. 

Practical applications of MTE include: integration of knowledge from
heterogeneous sources for object identification  \cite{deKorvin:93}, 
technical diagnosis under unreliable measuring devices  \cite{Durham:92}, 
medical applications: \cite{Gordon:90,Zarley:88b}.

An important example concerning difference between implications of bayesian
reasoning and reasoning within MTE can be found in \cite{Smets:94}. 

In spite of indicated merits, MTE experienced sharp criticism from many sides.
The basic line of criticism is connected with the relationship between the
belief function (the basic concept of MTE) and frequencies. 
 A number of attempts to interpret belief functions in terms of probabilities
have failed so far to produce a fully compatible interpretation with MTE - see
e.g. \cite{Kyburg:87,Halpern:92,Fagin:91} etc. Shafer
\cite{Shafer:90ijar} and Smets \cite{Smets:92}, in defense of MTE, dismissed
every attempt to interpret MTE frequentistically. Shafer stressed that
even modern (that meant bayesian) statistics is not frequentistic at all
(bayesian theory assigns subjective probabilities), hence frequencies be no
matter at all. Smets stated that domains of MTE applications are those where
"we are ignorant of the existence of probabilities", 
and warns that MTE is 
"not a 
model for poorly known probabilities" (\cite{Smets:92}, p.324). Smets states
further "Far too often, authors concentrate on the static component (how
beliefs are 
 allocated?) and discover many relations between TBM (transferable belief 
model of Smets) 
 and ULP (upper lower probability) models, inner and outer measures 
(Fagin and Halpern \cite{Fagin:89}), random sets (Nguyen \cite{Nguyen:78}), 
probabilities of provability 
 (Pearl \cite{Pearl:88}), probabilities of necessity (Ruspini 
\cite{Ruspini:86}) etc. But these authors 
usually do not explain or justify the dynamic component (how are beliefs 
updated?), that  is, how updating (conditioning) is to be handled (except in 
some cases by defining conditioning as a special case of combination). So I 
 (that is Smets) feel that these partial comparisons are incomplete, 
especially 
as all these interpretations lead to different updating rules." 
(\cite{Smets:92}, pp. 324-325).  Ironically, Smets gives later
in the same paper
 an example of
belief function ("hostile-Mother-Nature-Example") which may be clearly
considered as lower probability interpretation of belief function, just, at
further consideration, leading to very same pitfalls as approaches criticized
himself. 

Wasserman \cite{Wasserman:92ijar} strongly opposed claims of Shafer
\cite{Shafer:90ijar} about frequencies and bayesian theory. Wasserman pointed
out that the major success story of bayesian theory is  the
exchangeability theory of  de Finetti, which  treats frequency based
probabilities as a special case of bayesian belief. Hence frequencies, as
Wasserman claims, are inside the bayesian theory, but outside the
Mathematical Theory of Evidence.

This paper is intended to shed some light onto the dispute on   
relationship between MTE and frequencies. 
Section 2 introduces basic definitions of MTE.
Section 3 clarifies the problem
under consideration. Section 4  describes, how to interpret data in terms of
MTE 
belief functions. Section 5 describes a proposal of a model for structuring  n
of MTE belief functions. 
belief functions. Section 6 shows how to generate a random sample out of a
MTE model of reality. Section 7 describes how  \Bem{150} to 
derive MTE model from data. Section 8  is devoted to 
comparison of results of reasoning in MTE model and reasoning from data. 
Section 9 indicates open research problem not considered so far within the
presented interpretational framework of Mathematical Theory of Evidence.
\Bem{
4\sektion{A Frequentistic Interpretation of Belief Functions}%
5\section{MTE Models}
6\section{Generating Random Samples from MTE Models}
7\section{Extracting MTE Models from Data}
8\section{Reasoning in Shenoy-Shafer Axiomatic Framework versus Reasoning from
Data}
9\section{Discussion and Concluding Remarks}
}
\Sektion{2. BASIC DEFINITIONS OF MTE}

Let us first remind basic definitions of MTE:
 
\begin{df} 
Let $\Xi$ be a finite  set of elements called elementary events. 
Any subset of $\Xi$ be a composite event. $\Xi$ be called also the 
frame of discernment.\\
A basic probability assignment function is any function m:$2^\Xi  \rightarrow
[0, 1]$ such that  $$  \sum_{A \in 2^\Xi } |m(A)|=1, \ \ 
  m(\emptyset)=0, \ \  
\forall_{A \in 2^\Xi} \quad  0 \leq  \sum_{A \subseteq B} m(B)$$
($|.|$ - absolute value.\\
      
A belief function be defined as Bel:$2^\Xi \rightarrow [0,1]$ so that 
 $$Bel(A) = \sum_{B \subseteq A} m(B)$$
A plausibility function be Pl:$2^\Xi \rightarrow [ 0,1]$  with 
$$\forall_{A \in 2^\Xi} \  Pl(A) = 1-Bel(\Xi-A )$$
A commonality function be Q:$2^\Xi-\{\emptyset\} \rightarrow [0,1]$ with 
 $$\forall_{A \in 2^\Xi-\{\emptyset\}} \quad Q(A) = \sum_{A \subseteq B}
m(B)$$ \end{df}

Furthermore, a Rule of Combination of two Independent Belief Functions 
$Bel_1$,
 $Bel_2$ Over the Same Frame of Discernment (the so-called Dempster-Rule),
denoted 
    $$Bel_{E_1,E_2}=Bel_{E_1} \oplus Bel_{E_2}$$ 
 is defined as follows: :
$$m_{E_1,E_2}(A)=c \cdot  \sum_{B,C; A= B \cap C} m_{E_1}(B) \cdot 
m_{E_2}(C)$$ (c - constant normalizing the sum of $|m|$ to 1)

Furthermore, let the frame of discernment $\Xi$ be structured in that it is
identical to cross product of domains $\Xi_1$, $\Xi_2$, \dots $\Xi_n$ of n
discrete variables $X_1, X_2, \dots X_n$, which span the space $\Xi$. Let
$(x_1, x_2, \dots x_n)$ be a vector in the space spanned by the variables 
$X_1,
 ,  X_2, \dots X_n$. Its projection onto the subspace spanned by variables 
$X_{j_1}, X_{j_2}, \dots X_{j_k}$ ($j_1, j_2,\dots j_k$ distinct indices from
the set 1,2,\dots,n) is then the vector $(x_{j_1}, x_{j_2}, \dots x_{j_k})$. 
$(x_1, x_2, \dots x_n)$ is also called an extension of $(x_{j_1}, x_{j_2},
\dots x_{j_k})$. A projection of a set $A$ of such vectors is the set
$A ^{\downarrow X_{j_1}, X_{j_2}, \dots X_{j_k}}$ 
 of
projections of all individual vectors from A onto $X_{j_1}, X_{j_2}, \dots
X_{j_k}$. A is also called an extension of $A ^{\downarrow X_{j_1}, X_{j_2},
\dots X_{j_k}}$. A is called the vacuous extension of $A ^{\downarrow
X_{j_1},
 X_{j_2}, \dots X_{j_k}}$  iff A contains all possible extensions of each
individual vector in $A ^{\downarrow X_{j_1}, X_{j_2}, \dots X_{j_k}}$ .
The fact, that A is a vacuous extension of B onto space $X_1,X_2,\dots\,
X_n$ is denoted by $A=B ^{\uparrow X_1,X_2,\dots\,X_n}$
\begin{df}
Let m be a basic probability assignment function on the space of discernment
spanned by variables   $X_1,X_2,\dots\,X_n$. $m ^{\downarrow X_{j_1},
X_{j_2}, \dots X_{j_k}}$ is  called  the  projection  of  m  onto 
subspace spanned by
$X_{j_1}, X_{j_2}, \dots X_{j_k}$ iff 
$$m ^{\downarrow X_{j_1}, X_{j_2}, \dots X_{j_k}}(B)= c \cdot
\sum_{A; B=A  ^{\downarrow X_{j_1}, X_{j_2}, \dots X_{j_k}} } m(A)$$
(c - normalizing factor)
\end{df}
\begin{df}
Let m be a basic probability assignment function on the space of discernment
spanned by variables  $  X_{j_1},
X_{j_2}, \dots X_{j_k} $. $m ^{\uparrow X_1,X_2,\dots\,X_n}$ is called
the vacuous extension 
 of m onto superspace spanned by $X_1,X_2,\dots\,X_n$
iff 
$$m ^{\uparrow X_1, X_2, \dots X_n}(B ^{\uparrow X_1,X_2,\dots\,X_n})=m(B)$$

and $m ^{\uparrow X_1, X_2, \dots X_n}(A)=0$ for any other A. \\
We say that a belief function is vacuous iff $m(\Xi)=1$ and $m(A)=0$ for any A
different from $\Xi$.
\end{df}

Projections and vacuous extensions of Bel, Pl and Q functions are defined
with
respect to operations on m function. Notice that by convention if we want to
combine by Dempster rule two belief functions not sharing the frame of
discernment, we look for the closest common vacuous extension of their
frames of discernment without explicitly notifying it.

\begin{df} (See \cite{Shafer:90b}) Let B be a subset of $\Xi$, called 
evidence,
 $m_B$ be a basic probability assignment such that $m_B(B)=1$ and $m_B(A)=0$
for any A different from B. Then the conditional belief function $Bel(.||B)$
representing the belief function $Bel$ conditioned on evidence  B 
is defined
as: $Bel(.||B)=Bel \oplus Bel_B$. 
\end{df}

\Sektion{3. PROBLEM STATEMENT}

We notice easily, that if the function m is positive only for sets with
cardinality 1 then Bel is the plain finite discrete probability function. On
relationship between probability functions and bayesian rule consult e.g.
\cite{Halpern:92}. As m sums up to 1 on the set of sets of elementary events
we may be tempted to consider m as an ordinary probability function. But, as
cited previously from the work of Smets, this view contradicts the Dempster
rule of combination of independent Bels. 

In fact, probability functions give rise to some expectations, which cannot
be satisfied by a belief function.
Generally, we feel that a theory of real world is correct if we make initial 
observations,
 let run the intrinsic real world process and the theoretical process on  
this set of observations and then results of both processes coincide.
Probability theory matches this expectation. If we model the reality with a
joint probability distribution P() and set a selection condition to membership
in set B, start a real world process  generating independent events following
distribution P() and out of these events we strictly select those fitting
condition B, and then within those selected events we estimate a joint
probability distribution P'(), then P'() and P() are related (in the
limit) by the theoretical model of conditional distribution $P'(A)=P(A|B)$.
The major trouble with belief functions is that no such frequentist
interpretation  exists  for  them.  Models  criticized  by  Smets 
\cite{Smets:92}
and many other fail to define such an interpretation of initial data, the
process and the results as to fit Dempster rule of independent evidence
combination.  

Smets and Shafer propose a simple solution to this problem: do not look for
such real world processes at all. MTE shall exist as an nicely shaped
abstract
object in the space of philosophical thinking devoid of any  practical
meaning. 

We propose here another way out. We shall insist on looking for processes
meeting MTE requirements and rethink the type of process governing
probabilistic reasoning. 

\Sektion{4. A FREQUENTISTIC INTERPRETATION OF BELIEF FUNCTIONS}%

A detailed presentation of our interpretation is presented elsewhere
\cite{Klopotek:93b}. 

Essentially, we assume (like in papers of Nguyen \cite{Nguyen:78})  that each
object of the population has set-valued attribute(s). But unlike other
approaches we assume that objects are attached labels. Labels are subsets of
the frame of discernment $\Xi$  and  indicate  which  values  are 
permissible for
consideration for a given object. Then the belief function Bel(A) measures
for the set A the share of objects for which the intersection of the
attribute value of the object and of the label of this object (and NOT
solely the attribute value) is contained in A (m, Pl, and Q are easily derived
from Bel). 

The idea of a label reflects subjectiveness within the framework of MTE. It
happens in the real life that community attaches vicious labels to a man  
 remembering only his  weaknesses (which he surely has) and totally ignoring
his virtues (which he may possibly have). So the label of an MTE object  may
express a kind of irreversible prejudice, or attitude, or belief of a
community with respect to this object. 
We can also imagine a process of medical diagnosis. An iniatal  set of
hypotheses is  proposed. Then various medical tests are run which label the
patient with different sets of hypotheses, and the intersection of those
labels  and the initial finding is considered to be the actual ilness 
(though no single test may be possible to distinguish this single
ilness from all the other). 
Another type of labeling may be that
of actual loosing some properties an object had once upon a time. Someone
qualified today as a contemporary example  good father, may loose this
property in - say - 40 years (He may die by then or be just a good
grandfather). A match may loose its feature as being able to light a fire 
etc.

The importance of a label for the MTE object is operational. Let us assume
we run a deterministic MTE "selection" process $proc_B$ as follows: if the
intersection of B, the object's attribute value and its label is empty then
we reject the object from the population (negative selection, identical with a
probabilistic conditioning process). Otherwise we change the label of the
object to intersection of B and its pre-process label (this is unlike
probabilistic conditioning where we leave accepted objects unchanged). 
Let us denote by Bel' the belief function derived from the population obtained
after the process $Bel_B$. It is easily seen that then $Bel'=Bel \oplus
Bel_B=Bel(.||B)$. 

We may also consider a non-deterministic  family of such labeling processes 
$proc_{B_1}$, $proc_{B_2}$, \dots $proc_{B_k}$ where independently for each
object randomly one of the processes is chosen with probability $Pr(B_1)$,
$Pr(B_2)$, \dots $Pr(B_k)$ respectively. Let Bel denote the belief function
from frequencies in the initial population, Bel" the belief function after the
run of combined processes, and $Bel_{proc}$ shall denote such a belief
function
that for j=1,\dots,k $m_{proc}(B_k)=Pr(B_k)$. Then again is is easily 
demonstrated that the expected value of Bel" is equal $Bel" = Bel \oplus
Bel_{proc}$. 

Let us illustrate the idea with the following example. 

Let us assume that Mr.AX, Mr.BY, Mr.CZ and Mr.DT are major contributors of
(imaginary) Formal Theory of Formality (FTF). We shall evaluate their relative
contribution to the theory - just by measuring the number of publications. 
We collected 100 papers some of which were written  by  a  single 
author, some
by several of them. The overall statistics is presented in the subsequent
table: 
\begin{center}
\begin{tabular}{lc}
Author(s)  &  Number of publications\\
\hline
AX         &            5\\
BY         &           15\\
CZ         &            8\\
DT         &            2\\
AX,BY      &           24\\
AX,CZ      &           11\\
AX,BY,CZ   &           20\\
AX,BY,DT   &            9\\
AX,BY,CZ,DT&            6\\
\hline
TOTAL      &          100\\
\end{tabular}
\begin{tabular}{c|ccc}
fun/arg & $\{AX\}$ & $\{BY\}$ & $\{AX,BY\}$\\
\hline
m       &  0.05    & 0.15     & 0.24\\
Bel     &  0.05    & 0.15     & 0.44\\
Pl      &  0.75    & 0.74     & 0.90\\
Q       &  0.75    & 0.74     & 0.59\\
\end{tabular}
\end{center} 

The neighboring table shows the values of MTE measures for three sets of
authors consisting of : (1) only AX, (2)only BY and (3) both AX and BY.

Let us assume that we obtained a "hint" from a "friendly person" that Mr.DT is
in fact a fictive author of FTF. This information ("evidence")  may be
captured by the belief function  $Bel_1$ such that 
$m_1(\{AX,BY,CZ\})=1$. How to use this hint ? We may take the papers
we collected, one by one, and delete Mr.DZ from the list of authors of the
paper, and if no author is left, we throw the paper away. After such an
operation we obtain the new "statistics" of contributions:
\begin{center}
\begin{tabular}{lc}
Author(s)  &  Number of papers:\\
\hline
AX         & 5\\
BY         &15\\
CZ         & 8\\
AX,BY      &33\\
AX,CZ      &11\\
AX,BY,CZ   &26\\
\hline
TOTAL      &98\\
\end{tabular} %
\begin{tabular}{c|ccc}
fun/arg & $\{AX\}$ & $\{BY\}$ & $\{AX,BY\}$\\
\hline
m'      &   5/98   & 15/98    & 33/98\\
Bel'    &   5/98   & 15/98    & 53/98      \\
Pl'     &  75/98   & 74/98    & 90/98      \\
Q'      &  75/98   & 74/98    & 59/98      \\
\end{tabular}
\end{center}
Bel' be the new belief function for the new population. It is easily seen that
$Bel' = Bel \oplus Bel_1$. 

Let another, independent friendly person give us another hint that also CZ is
a fictive contributor to FTF. But let us trust this person only to 70~\%. 
This fact should be represented by the belief function  $Bel_2$ such that
$m_2(\{AX,BY,DT\})=0.7$ and  $m_2(\{AX,BY,CZ,DT\})=0.3$.

Let us consume this new "evidence" as follows: we take papers (after last
operation) one by one, throw a properly biased coin (70\% heads, 30 \% tails),
toss it and on heads we delete CZ from the list of authors (and throw the
paper away if no other author is left) and on tails we accept the paper as is.
The expected value over "statistics" of a process like this is visible 
below.\\
\begin{center}
\begin{tabular}{lc}
Author(s)   &   Number of papers:\\
\hline
AX         & 8.3\\
BY         &15\\
CZ         & 5.6\\
AX,BY      &40.8\\
AX,CZ      & 7.7\\
AX,BY,CZ   &18.2\\
\hline
TOTAL      &95.6\\
\end{tabular} %
\end{center}
Let   $Bel"$ denote the belief function after the second process. Provably its
expected value may be calculated as $Bel"=Bel' \oplus Bel_2$. 

\Sektion{5. MTE MODELS}

It is usually (next to) impossible to represent directly a (joint) probability
distribution in a larger number of variables. E.g. 15 four-valued variables
would require 1 Giga floating point cells. With MTE belief functions it is
even worse. A direct representation of a belief function in 15 
variables with four-valued frame of discernment each would require 1 billion
Giga floating point cells. Therefore in both domains alternative
representations are looked for. Within probability domain so-called bayesian
networks are used \cite{Geiger:90}, exploiting conditional independences,
where the joint  probability  distribution  is  represented  as  a 
combination of
conditional probabilities of variables $X_i$ on their direct causes $X_{\pi
(i)}$:
$$P(x_1,...,x_n) = \prod_{i=1}^{n} P(x_i | x_{\pi (i)})$$

Within the MTE literature another kind of "belief networks" is used, due i.e.
to Shenoy and Shafer \cite{Shenoy:90}. It is so-called factorization along a
hypergraph:
$$Bel = Bel_1 \oplus Bel_2 \oplus \dots \oplus Bel_m$$
where no requirements are put on the  form of components $Bel_i$. This
approach has been criticized recently by Cano et al. \cite{Cano:93} as -
contrary to Pearl's bayesian networks - these MTE belief networks don't
reflect (conditional) independences among variables.
Shafer's conditional belief function cannot be candidate for a factor in
belief function factorization.
Therefore they proposed a
definition of (so-called apriorical) conditionality  within MTE saying that
$Bel$ is a conditional belief function conditioned on variables $X_1,\dots
X_k$ iff $Bel ^{\downarrow X_1,\dots,X_k}$  is a vacuous belief function. They
require the factorization of a belief function to be in terms of such
conditional belief functions. However, this definition is counterproductive as
for the special case of belief function being a bayesian distribution the
whole bayesian network may collapse into a single node of Cano's belief
network. 

We proposed therefore another definition of apriorical conditional belief
function in \cite{Klopotek:93f} and demonstrated there that Shenoy/Shafer
factorization cannot be simpler than that into a belief network of ours.
Apriorical conditional belief function $Bel ^{|X_1,\dots,X_k}$ reflecting
conditioning of $Bel$ on $X1,\dots,X_k$ is any pseudo-belief function
satisfying $$Bel =  Bel ^{|X_1,\dots,X_k} \oplus  Bel ^{\downarrow X_1,\dots,
 X_k}$$ (A pseudo-belief functions allows m's to be negative, but only so that
respective Q's remain non-negative). We then define a belief network
factorization of  $Bel$  with  frame  of  discernment  spanned  by 
variables $X_1,
\dots, X_n$ as 
$$Bel  = \bigoplus_{i=1}^{n}Bel ^{\downarrow \{X_i\} \cup X_{\pi (i)} |  
X_{\pi (i)} } $$

\Sektion{6. GENERATING RANDOM SAMPLES FROM MTE MODELS}

Generating a random sample from the bayesian network is a relatively simple
task. We have to consider an ordering of variables compatible with the partial
order imposed by the network. To generate a single object,  we take variable
by variable following this
ordering, look at values  $x_{\pi(i)}$ already assigned to direct causes
$X_{\pi(i)}$ of $X_i$
 and then we select randomly one of the values $x_{i1},x_{i2},\dots,x_{in_i}$
taken by variable $X_i$ with probabilities        proportional to 
 $P(x_{i1}|x_{\pi(i)}),P(x_{i2}|x_{\pi(i)}),\dots,P(x_{in_i}|x_{\pi(i)})$ 
That is, in a single pass we can generate one sample object. (All random
selections should be independent of one another). 

It is not that easy with belief function models in general. To generate a
single sample
object we have to assign it first the frame of discernment set $\Xi$, calling
this set $O_0$. Then we take one after the other component $Bel_i$ function
(ordering may be
anyhow), eventually make the vacuous extension  $Bel_i 
^{\uparrow}$  of it onto
the frame of discernment. Then we select randomly a set $A_j\subseteq\Xi$
according to probability distribution proportional to $|m_i
^{\uparrow}(A_j)|$.
If at the time the object had the value $O_{i-1}$, then we assign it a new
value $O_i=O_{i-1} \cap A_j$. Should $O_i$ be an empty set, then we terminate
the pass without generating an object (a failure). That is, within a single
pass we may or may not generate a sample object. \\
With pseudobelief functions the situation is even worse. The object is within
each step assigned a set and a mark + ore -. If the $m_i(A_j)$ of selected
$A_j$ is positive, then the mark is left unchanged, if it is negative, then
the mark is inverted. If within the sample there are two identical objects
with respect to the finally assigned set, but with opposed marks, then they
are canceled out. After completion of sample generation process all remaining
objects with marks '-' are canceled.

However, if the model is in terms of a belief network as described in 
 \cite{Klopotek:93f} (see section
5),    then the problem of sample generation reduces to probabilistic case
(one
sample object per single pass), however one has to keep track of '+'/'-' marks
as indicated above.
\Sektion{7. EXTRACTING MTE MODELS FROM DATA}

One of major advantages of the belief network model proposed by us for MTE is
its capability to connect statistical independence from data with d-separation
within the underlying directed acyclic graph of the belief network (see
\cite{Klopotek:93d} for more on this point). d-separation property
\cite{Geiger:90} means possibility of derivation of conditional independence
from graphical structure of dependencies. Several algorithms have been
proposed for derivation of belief network structure for MTE (see
\cite{Klopotek:93b,Klopotek:93f,Klopotek:93d}).

 Within the
framework of DS-JACEK system \cite{Michalewicz:93b}
still another
algorithm has been implemented. It is a version of CI algorithm of Spirtes et
al. (see \cite{Spirtes:93}. CI algorithm is known of capability to derive
causal models under causal insufficiency. It has been adopted for generation
of bayesian networks \cite{Klopotek:93g}, and later for MTE belief
networks \cite{Klopotek:93h}.
The essential adaptation step for MTE is substitution of bayesian conditional
independence test with MTE apriorical conditional independence test. If
we test conditional independence of variables $X$ and $Y$ on the set of
variables $Z$, then we have to compare empirical distribution $Bel
^{\downarrow X,Y,Z}$ with $Bel ^{\downarrow X,Z|Z} \oplus Bel ^{\downarrow Y,
Z|Z} \oplus Bel ^{\downarrow Z}$. The traditional $\chi ^2$ statistics is
computed (treating  the latter distribution as expected one).  If the
hypothesis of equality is rejected on significance level $\alpha=0.05$ then
X and Y are considered dependent, otherwise independent. 

\Sektion{8. REASONING IN SHENOY-SHAFER AXIOMATIC FRAMEWORK VERSUS REASONING
FROM DATA}

The crucial point of the whole effort of developing data model for MTE was to
provide a reference point for MTE reasoning engines.

An MTE reasoning engine calculates (Shafer's) conditional marginal
distributions from an MTE model and a set of observations. Several such
engines have been developed (see \cite{Shenoy:90} for further references, see
also \cite{Cano:93}). Shenoy/Shafer method of local computations for MTE,
described e.g. in \cite{Shenoy:90} 
works with hypergraph factorization models of belief functions. It
has been implemented in the reasoning
engine of DS-JACEK \cite{Michalewicz:93b}.

Conditioning for probabilistic databases is achieved in a simple manner. It is
sufficient to impose a filter (filtering expression, or select expression) and
the calculation of marginals can be done on the physically same dataset just
evaluating truth value of filtering expression. 

It is not that simple for
MTE databases.
Neither are MTE constraints deterministic (they are usually non-deterministic)
nor is conditioning just the matter of selection.  One has to prepare the
physical copy of the database. Then take each record case-by-case and apply
the same procedure as when generating a random sample with two differences:
we do not start with assignment of $\Xi$ to the object, but rather start with
its currently assigned set of values. Then we do not use Bels describing the
model but rather the ones describing constraints (observations). So the record
may be either discarded from the database or retained but changed (Due to
possibilities of changes we just require that a physical copy of the original
database is to be prepared for the process). Then marginals over the modified
copy are just results of MTE reasoning from data.

It is a very important issue to possess two different mechanisms for
MTE reasoning: one from model and one from data. We may have diverse testing
conditions where we may utilize them in suitable way. On the one hand we may
have a model of reality available and want to develop/test a new/old MTE
reasoning engine. It is usually to hard to verify results of MTE reasoning for
models with more than 10 variables by hand. And usually the implementation of
sample generator in combination with sample constraint imposer is much simpler
than that of a fine (time and/or space saving) model-based MTE reasoning
engine. So we can generate a random sample from the model, run reasoning both
via sample constraint imposer and via the developed knowledge-based reasoning
engine and then compare resulting (aposterioric) conditional belief functions
from data and from the model.\\
The other setting may be that we have data available and want to develop the
proper model of reality. We may generate model from data either by automatic
learning program, or in interactive manner or by "enlightened guess". Then
we can compare under various constrains the results of reasoning from this
model and the data and eventually reject or adjust the model, change/improve
model generation algorithms etc..\\
Last not least, we may have been provided with a model once upon the time and 
it was not until later that we collected a sufficient amount of data to verify
its thorough or partial validity.  

\Sektion{9. DISCUSSION AND CONCLUDING REMARKS}

The frequentistic interpretation of Mathematical Theory of Evidence described
in this paper has been fully implemented: major part of it within a
knowledge-base development-and-test system DS-JACEK and partially within the
test-bed for this system. For sparsely connected networks of up to 15
dempsterian variables,  databases with  up  to  5,000  cases  were 
randomly
generated and thereafter in most cases successfully reconstructed using
methodology described above. Cases of failure turned out usually to provide
alternative but functionally equivalent structure. Run-times on an PC i486,
66MHz in extreme cases didn't exceed 10 minutes.

Generation of random dempsterian samples served also as a test-bed for the
reasoning machine of DS-JACEK. Following of test examples with more than 10
variables by hand may prove not feasible. Hence for larger MTE models random
samples of 10,000 cases and more were generated and then results of
knowledge-based DS-JACEK reasoning machine were compared with marginals from a
sample-based test-bed  reasoning machine. Comparisons of this form proved to
be quite useful for program development.

With this presentation main application problems of MTE seem to be overcome.
Processes changing data in a coherent way (reducing the space of possibilities
for each individual) like those of stepwise medical diagnosis, or destructive
processes have been identified as a class of processes which can be modeled
using Mathematical Theory of Evidence. Belief functions may be calculated
from data, verified against data and may serve as source for random generation
of data. Independence of dempsterian-shaferian variables acquired statistical
meaning. A new category of MTE belief networks has been defined paralleling
bayesian belief networks in their capability to capture qualitatively
independence and conditional independence in form of a directed acyclic
graphs. Algorithms to decompose a data grounded belief function into a belief
network have been elaborated.

A number of questions are still left open. :
\begin{itemize}
\item how to interpret (if at all) pseudo-belief functions (that is with
negative masses in basic probability assignment),
\item how to obtain unbiased  estimates of apriorical conditional belief
functions weighing belief network nodes,
\item how to accomplish statistically optimal approximations of belief
functions by belief networks from noisy data,
\item what should be optimal independence and conditional independence tests
for belief network models.
\end{itemize}

These questions along with related issues are subject of further research.
 It
is due to the work presented in this paper that questions of statistical
estimations within the MTE may be posed at all.

 Previous work, done among
other by Shafer \cite{Shafer:90b} 
by Kyburg \cite{Kyburg:87}, Halpern and Fagin \cite{Halpern:92}, Fagin
and Halpern \cite{Fagin:89,Fagin:91}, Pearl \cite{Pearl:90},
Grzymala-Busse
\cite{Grzymala:91}, 
Nguyen \cite{Nguyen:78}, as well as critical evaluations of
these approaches e.g. by Smets \cite{Smets:92}, contributed to clarification
of similarities and dissimilarities between frequentistic approaches to
probability and to MTE. The major contribution of this paper is to pinpoint
the essential difference between probabilistic reasoning and the reasoning
within MTE. Probabilistic reasoning can be understood - in terms of objects
of the population - as a selection of objects with some properties. Whereas
MTE reasoning means selection of objects combined with destruction of
some of their properties. So, objects subject to a "probabilistic
process" are the same before and after completion of the process.
Whereas objects subject to an MTE process are changed, deformed after
completion of the process. I think this explains difficulties
encountered by alternative frequentistic interpretations of MTE
presented in the literature. This fundamental insight allowed for
enhancement of MTE with many features reserved so far for probabilistic
models only (like representation in terms of a belief network,
generation of random samples from belief function models, identification of
belief networks from data, comparison of model-based and data-based
reasoning  etc.).  

\small 

\newcommand{\LitStelle}[2]{\bibitem{#1}}

\newcommand{\ReadingsIn}{G. Shafer, J. Pearl eds: {\it Readings in Uncertain 
Reasoning}, (ISBN 1-55860-125-2, 
Morgan Kaufmann Publishers Inc., San Mateo, California, 1990)}

\end{document}